# Analysis of Rainfall Variability and Water Extent of Selected Hydropower Reservoir Using Google Earth Engine (GEE): A Case Study from Two Tropical Countries, Sri Lanka and Vietnam

Punsisi Rajakaruna, Surajit Ghosh[1], Bunyod Holmatov

International Water Management Institute

**Abstract**

This study presents a comprehensive remote sensing analysis of rainfall patterns and selected hydropower reservoir water extent in two tropical monsoon countries, Vietnam and Sri Lanka. The aim is to understand the relationship between remotely sensed rainfall data and the dynamic changes (monthly) in reservoir water extent. The analysis utilizes high-resolution optical imagery and Sentinel-1 Synthetic Aperture Radar (SAR) data to observe and monitor water bodies during different weather conditions, especially during the monsoon season. The average annual rainfall for both countries is determined, and spatiotemporal variations in monthly average rainfall are examined at regional and reservoir basin levels using the Climate Hazards Group InfraRed Precipitation with Station (CHIRPS) dataset from 1981 to 2022. Water extents are derived for selected reservoirs using Sentinel-1 SAR Ground Range Detected (GRD) images in Vietnam and Sri Lanka from 2017 to 2022. The images are pre-processed and corrected using terrain correction and refined Lee filter. An automated thresholding algorithm, OTSU, distinguishes water and land, taking advantage of both VV and VH polarization data. The connected pixel count threshold is applied to enhance result accuracy. The results indicate a clear relationship between rainfall patterns and reservoir water extent, with increased precipitation during the monsoon season leading to higher water extents in the later months. This study contributes to understanding how rainfall variability impacts reservoir water resources in tropical monsoon regions. The preliminary findings can inform water resource management strategies and support these countries' decision-making processes related to hydropower generation, flood management, and irrigation.

*Keywords - Reservoir Extents, Hydroelectric reservoir, Rainfall, Sentinel-1, GEE*

1. **INTRODUCTION**

Tropical monsoon countries like Vietnam and Sri Lanka, which have substantial weather exposure, could be ideal for monitoring the variation of water bodies concerning the changing weather [1]. Global climate change influences the long-term trend of climate variables such as rainfall, temperature, and wind. It is essential to analyze the long-term trend in rainfall patterns when a study of change in the reservoir water extent is conducted for a tropical region subjecting to periodic monsoons [2].

Vietnam uses hydroelectricity to fulfil a significant part of its electric power demand since the country has many reservoirs built for hydropower generation, flood management and irrigation [3]. Among them, 2900 reservoirs are contributing towards hydropower and irrigation while the total reservoir capacity of the country is 28 billion $m^3$. According to the Department of Water Resource Management in Vietnam, the availability of total per capita renewable water resources is gradually declining and is predicted to be only 3100 $m^3$ by 2025 as it depends on the upstream countries [4]. Identifying water storage capacity is a noteworthy problem in the country that needs to be addressed.

---

[1] Contact: s.ghosh@cgiar.org

Sri Lanka is influenced by two monsoons, southwest and northeast, and the seasons in Sri Lanka consist of these two monsoons and two inter monsoons. The direction of both monsoons impacts the rainfall pattern of the country. The southwest monsoon of Sri Lanka happens from May to September and is more impactful for the country's southwest region. The northeast monsoon usually starts in December and continues till February. It is creating more influence in the northeast region of the country [5]. Victoria is the largest power station that contributes electricity generation with a capacity of 210 MW. Kotmale provides a capacity of 201 MW. Both reservoirs are in the Mahaweli cascade. Samanalawewa is also a leading hydropower plant contributing to the countries' hydroelectricity with a capacity of 124 MW located in the Walawe cascade [6,7,8].

EO datasets are widely used to monitor the reservoir extent. Researchers can use high-resolution or moderate-resolution multi-spectral optical imagery from satellites with high revisit capacity for continuous monitoring, depending on their needs. Radar imagery can be used to overcome multiple problems with the use of optical imagery to map water bodies, especially during the monsoon season. Since the radar sensors are active sensors which can monitor the earth during day and night and have the capability of cloud penetration, they are highly recommended to be used in change detection during the rainy season [9].

The technological development of remote sensing instruments and advanced algorithms developed precipitation products such as CHIRPS (Climate Hazards Group Infrared Precipitation with Station) data [10]. Google Earth Engine (GEE) is a cloud computing platform introduced by Google in 2010 to conduct geospatial analysis with big data. Before GEE, Amazon Web Services and Microsoft Azure were introduced to the world to work with geospatial data. However, the benefit of GEE is that it supports more data types and is freely available. Many currently use GEE for their spatial analyses for the research and studies on climate change, agriculture, land use, disaster management, climate change impact, etc. The built-in functions, ability to export the desired outputs to be used in other applications like ArcGIS, global datasets created for specific use and interactive programming environment enhance the usability of GEE for geospatial analyses [11]. Combining remotely sensed precipitation and Sentinel-1 data to observe the change in the extent of water bodies is the main objective of this research. We used CHIRPS daily precipitation data and Sentinel 1 GRD satellite imagery to detect the monthly changes in the extent of three hydroelectric reservoirs in Vietnam and three hydroelectric reservoirs in Sri Lanka from 2017 to 2022. The present study aims to identify the relationship between the reservoir water extent using remotely sensed data and how the rainfall trend of the relevant areas has changed from 1981 – 2022.

## 2. STUDY AREA

The study focuses on two tropical countries, Vietnam and Sri Lanka. Vietnam experiences high temperatures and humidity throughout the year as it is in the tropical and temporal zones. The region is affected by the Southwest Monsoon from May to November, while the annual rainfall lies between 700 - 5000 mm. Flood duration in the country is about six months, from July to December, and extreme flood events occur between late September to mid-October[12]. The country has 2900 reservoirs having a capacity of 28 billion $m^3$, contributing to the hydropower and irrigation requirements of the country [4]. We have selected Tri An, Yali and Thac Ba reservoirs, which contribute to the hydroelectricity generation in the country (**Fig 1**).

Sri Lanka is in the tropical monsoon climate zone with a land area of 65,610 $km^2$. Therefore, the country experiences two monsoon periods: southwest monsoon (May- September) and northeast monsoon (December- February). The average annual rainfall over the country depends on the country's spatial location; the highest rainfall is expected at the central highlands, which exceeds a value of 5000mm. The southeast lowlands have a rainfall of 1000mm [5]. Most of the hydropower stations within the country are operated under CEB (Ceylon Electricity Board), and the output of these power plants has been heavily impacted by the recent climate change and deviations in water use due to that. Mahaweli Cascade contains the three largest hydropower plants in the country and contributes to the total hydropower by generating over 800 MW [13]. In the present study, we selected Victoria, Kotmale and Samanala Wewa reservoirs from Sri Lanka (**Fig 1**).

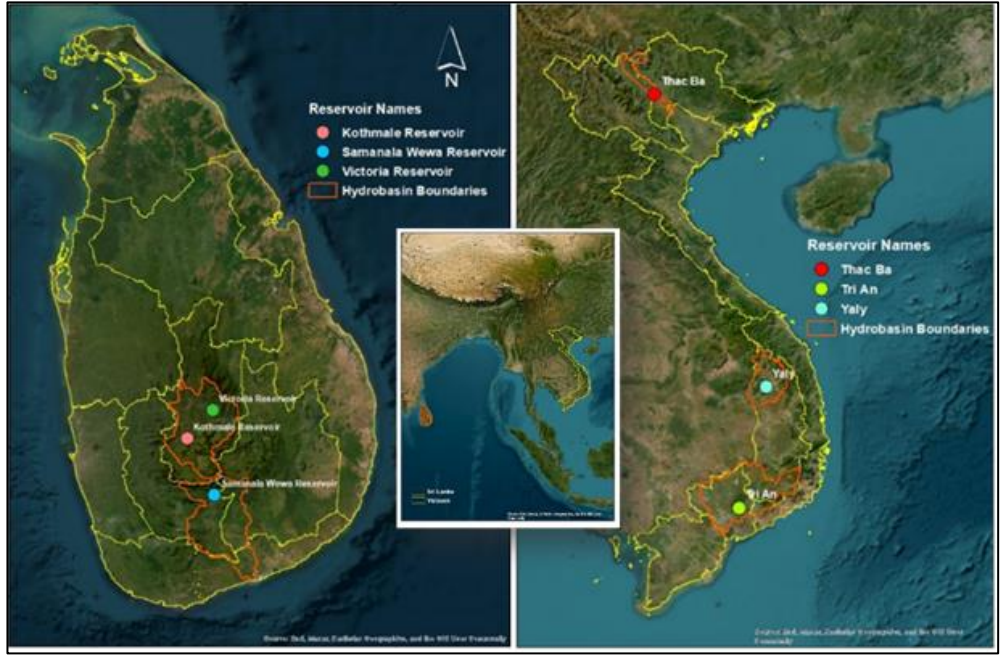

**Fig 1**. Hydrobasins of selected reservoirs

3. **METHODOLOGY**

The methodology (**Fig 2**) contains the comprehensive workflow of rainfall analysis and the reservoir extent calculation. The first phase describes the rainfall analysis, while the second phase explains the monthly water extent of the hydroelectric reservoirs.

*A. Remote Sensing Datasets*

   *a) Sentinel 1 SAR GRD Images*: The Sentinel 1 Ground Range Detected (GRD) dataset consists of daily updating image collection since 2014. Each Sentinel 1 scene includes three resolutions, four band combinations and three instrument modes and is pre-processed by following thermal noise removal, radiometric calibration, and terrain correction. The study used imagery captured with 10 m resolution and interferometric wide swath (IW) mode to derive the water extent of the existing reservoirs. The pre-processed data were acquired using an analysis-ready data cube with all the corrections applied to sentinel 1 SAR data. This ARD concept allows for rapid and easy use of complex datasets like SAR data. ARD helps the user avoid all the pre-processing complexities and concentrate on the use and analysis of the data.

   *b) CHIRPS Satellite Precipitation Dataset:* Climate Hazards Group InfraRed Precipitation with Station data (CHIRPS) and U.S. Geological Survey developed this IR-based precipitation dataset from 1981 to the present. The CHIRPS data have a high spatial resolution of 0.050 degrees. The gridded rainfall time-series dataset was formed by integrating satellite estimations and gauge observations with global climatology data. Here, we used the CHIRPS Daily (Version 2.0 Final) dataset to determine the rainfall trend of the study area.

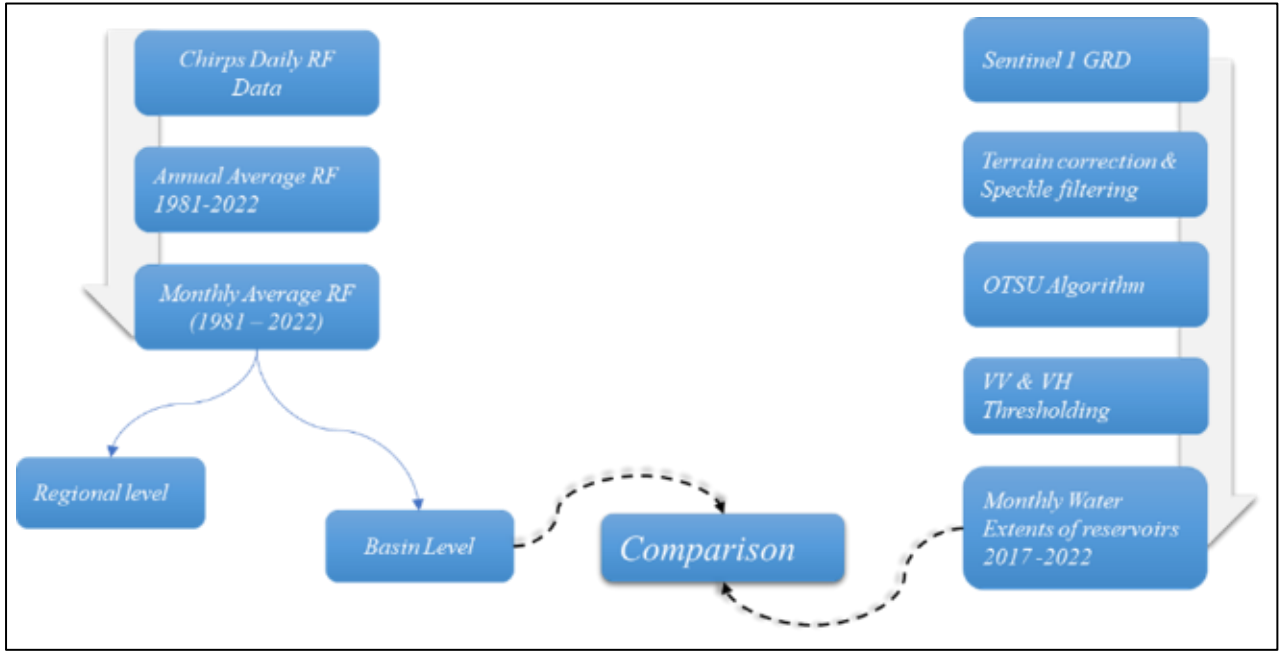

**Fig 2**. Methodology of the study

*B. Rainfall Analysis and Water Extent Mapping*

The CHIRPS Daily Precipitation Data was imported from the GEE data catalogue from 1981 – 2022. The Average annual rainfall maps for the countries were generated. Then, the regional/provincial level monthly average rainfall was calculated in GEE to understand the spatiotemporal variation. The monthly average rainfall values for each reservoir basin were also calculated for the same period. The reservoir water extents were measured with SAR data and the OTSU algorithm for water extent mapping. Sentinel 1 images with IW mode and 10 m resolution were loaded and corrected using terrain correction and refined Lee filter. Then, the corrected imagery was uploaded to assets for further use. Both VV and VH polarization were considered. Then, the VV and VH threshold values were obtained using the OTSU algorithm. OTSU thresholding algorithm is used to automatically detect an optimal threshold using a grey-level histogram based on the distribution of pixel values. The equation of the OTSU thresholding algorithm is,

$$\sigma^2(t) = P_w(t) \times \sigma_w^2(t) + P_{nw}(t) \times \sigma_{nw}^2(t) \quad (1)$$

Here, the $\sigma$ is the weighted sum of the variance between the water and non-water classes. Moreover, the $P_w$, $\sigma_w$ and $P_{nw}$ and $\sigma_{nw}$ is the variance and the probabilities of water (w) and non-water (nw) classes.

Since the backscatter intensity of the features is different in different conditions, an automatic OTSU algorithm is used to determine the dynamic threshold values for both VV and VH bands that separate water and land [14]. The connected pixel count threshold was also used to increase the accuracy of the result by removing the random pixels detected as water. These two steps were implemented for each corrected image and derived the water extents during each month for every reservoir from 2017 to 2022. Then, the water extent was calculated in square kilometres.

## 4. RESULTS AND DISCUSSION

The above line graph represents the annual average rainfall of the two countries. Both countries have an annual average rainfall of over 1500 mm (**Fig 3**).

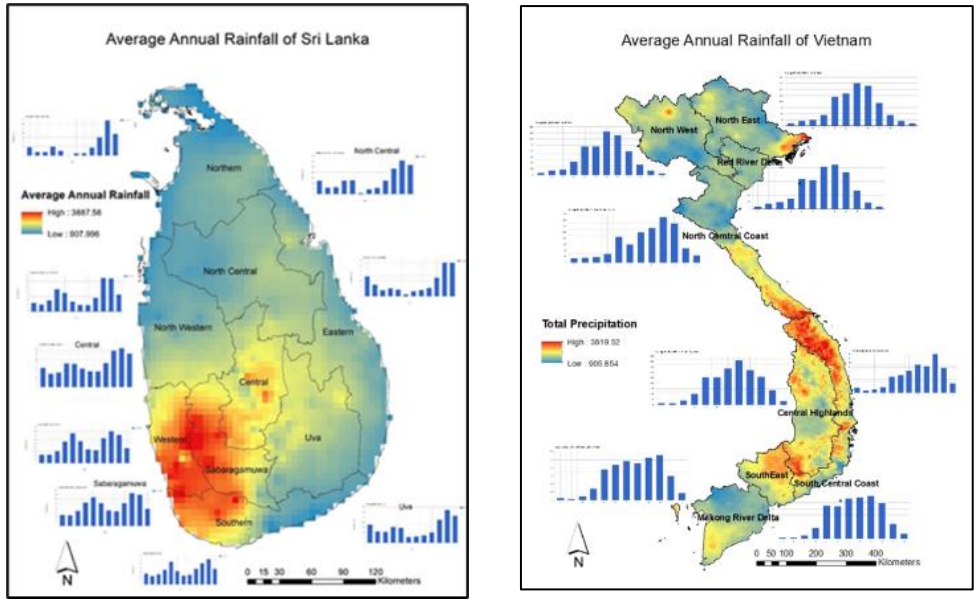

**Fig 3**. Regional Level Average Annual Rainfall Map and Monthly Average Rainfall Graphs of Sri Lanka and Vietnam

Several climate and environmental changes influence the change in reservoir water extent. From all of them, rainfall trend is also an influencing factor for dynamic reservoir water extents. Analyzing the relevant rainfall pattern of the relevant region or the hydro basin is also important to understand the variability of reservoir water extent.

Box and Whisker plot is a statistical visualization method that provides a synopsis of the distribution of the continuous variable. It is widely used to compare the distribution of variables within several groups and to determine the data spread. The key advantage of a box plot is that it indicates the data outliers. Figure 4 represents the box and whisk diagram of rainfall in Vietnam. According to that plot, the precipitation range and the values during the monsoon period are higher than in the dry season. The Box and Whisk diagrams for three reservoir extents in Vietnam and Sri Lanka are represented in **Figure 5** and **Figure 6**, respectively. The three reservoirs are showing a rise in reservoir extent after July. After October, the water extent shows only tiny deviations for those three reservoirs.

**Figure 4** also describes the box-and-whisker plots for rainfall data of Sri Lanka. The rainfall plot shows an increase in the precipitation values after September. In the reservoir extent plots for Sri Lanka, Kotmale shows a significant variation of extents than Victoria and Samanala Wewa. All three reservoirs are showing an increment of water extent after May.

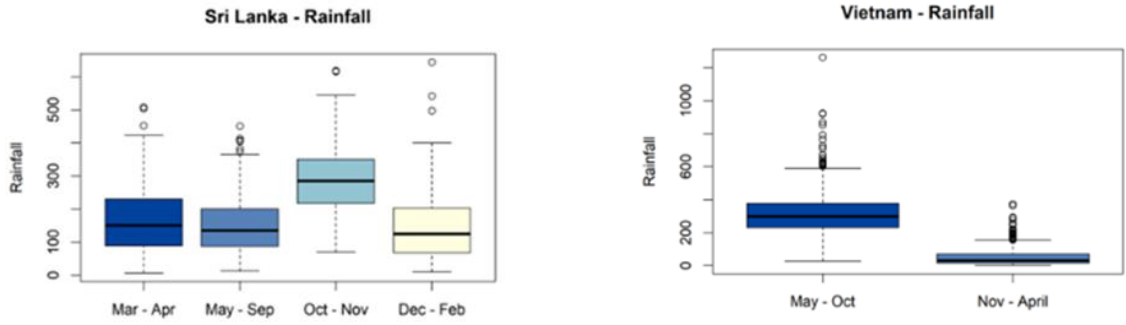

**Fig 4**. Box and Whisk Diagrams of Rainfall (top-Vietnam, bottom - Sri Lanka)

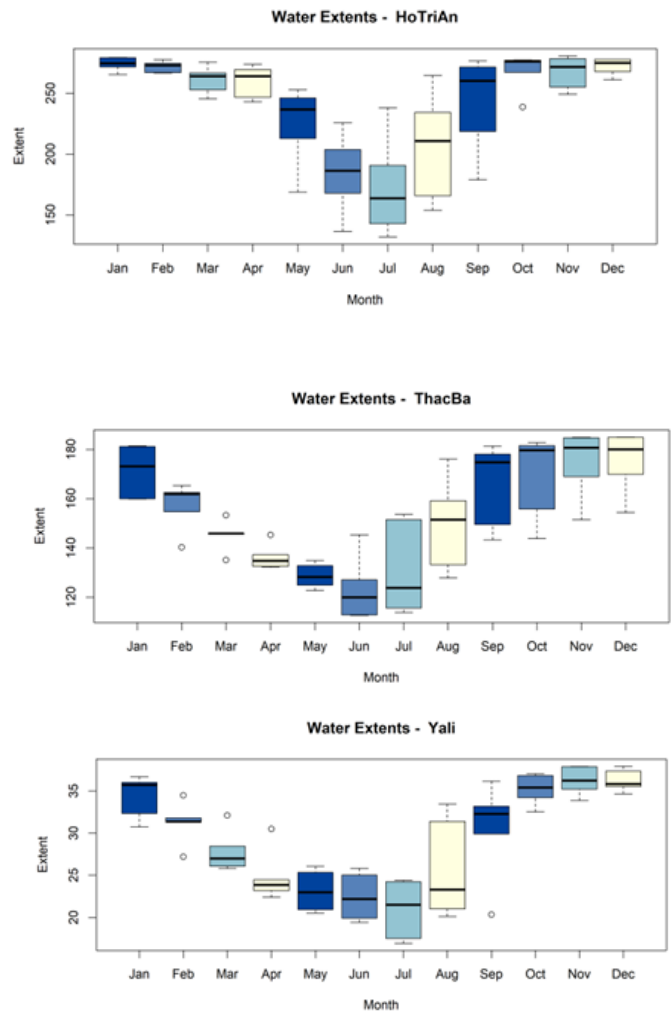

**Fig 5**. Box Whisk Diagrams for Water Extents Vietnam

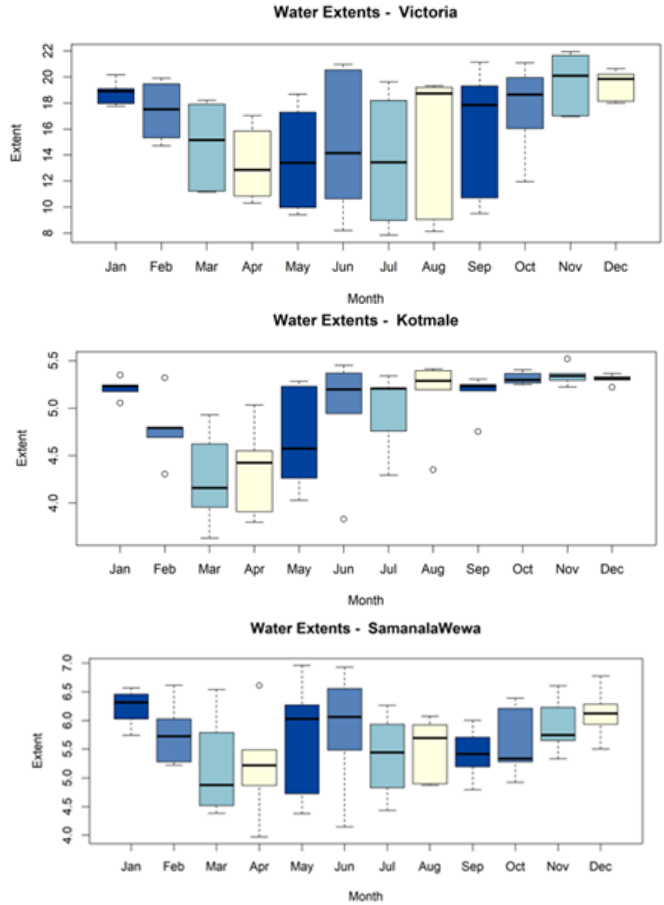

**Fig 6**. Box and Whisk Diagrams of Water Extents Sri Lanka

Monsoon precipitation is considered a primary source of water accumulation in the reservoir (**Fig 7**), which is further used for irrigation and other services. The minimum and maximum water extent of three selected reservoirs of 2022 in Sri Lanka were mapped. The minimum extent is 10.85 km$^2$, estimated in April, while the maximum area is 21.23 km$^2$, reported in November. In January, the rainfall over the hydro basin is at its minimum value; in October, it is at its maximum value. The water extent of the Victoria Reservoir in January is 17.97 km$^2$ and 19.84 km$^2$ in October.

Kotmale Reservoir is also located in Mahaweli Basin. Therefore, it also receives the minimum rainfall in January and maximum rainfall in October. The maximum water extent of the Kotmale reservoir in 2022 is reported in August, and the minimum water extent is reported in March. The respective minimum and maximum extents are 3.63 km$^2$ and 5.41 km$^2$. The water extent of the reservoir in January 2022 was 5.171 km$^2$, while the extent was 5.28 km$^2$ in October, where the minimum and maximum rainfall were reported (**Fig 8**).

Samanalawewa reservoir in the Walawe basin also received the maximum rainfall for 2022 in October, while minimum rainfall was received in January. The maximum water extent of 6.00 km$^2$ of Samanalawewa was found in June 2022, and the minimum water extent of 4.55 km$^2$ was estimated in March. The water extent in the month with the lowest rainfall is 5.74 km$^{2,}$ and 4.92 km$^2$ in October, the month with the highest rainfall for 2022.

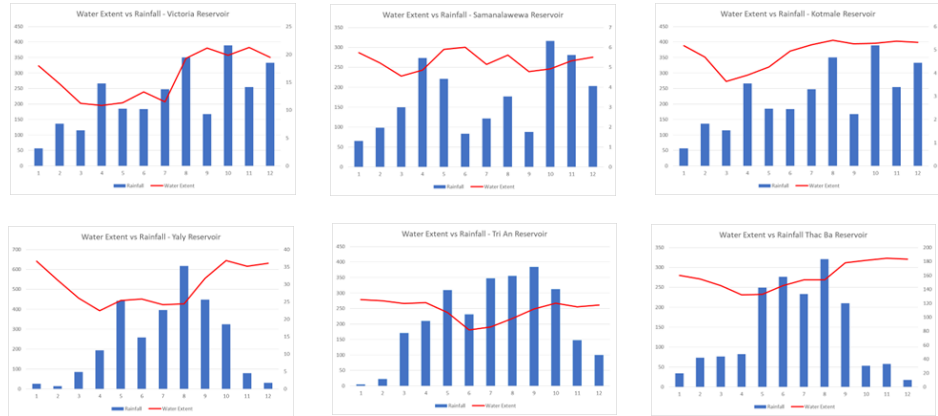

**Fig 7**. Water Extent and Rainfall Graph (first row – Sri Lanka, second row- Vietnam)

The maximum rainfall in 2022 for the Chay hydro basin where the Thac Ba is located is received in August, and the minimum rainfall value is reported in December. The minimum water extent of Thac Ba was reported in April as a value of 132.86 km$^2$, and the maximum water extent of 187.44 km$^2$ was identified in November. The water extent of Thac Ba reservoir in August with maximum rainfall was 155.25 km$^2$, while it was 186.37 km$^2$ in December, which had minimum rainfall. Yaly Reservoir, located in the Se San basin, received the maximum rainfall in August 2022, while it was a minimum in February.

The maximum water extent of Yaly occurred in October, while April reported the minimum water extent in 2022. The minimum and maximum water extent values are 22.40 km$^2$ and 36.82 km$^2$, respectively. The water extent of Yaly in August is 24.40 km$^2$ and 31.28 km$^2$ in February. The maximum water extent of Tri An was determined in January, while the minimum water extent value was in June for the year 2022. The minimum extent is 199.40 km$^2$, while the maximum water extent is 286.01 km$^2$. Dong Nai hydro basin, where the Tri An is situated, received the maximum rainfall in September and the minimum in January for 2022. The reservoir extent of Tri An in September 2022 is 255 km$^2$ (**Fig 9**).

According to the graphical representations of rainfall and water extents, it can be seen that the reservoir extents increase gradually sometime after the rainfall increases. Similarly, during the dry seasons, the reservoir extents gradually decrease, and these deviations in the reservoir extents are not instant. Human interventions for reservoir management and changes in inflow and dam operations also impact the deviation of reservoir water extents [15, 16]. The excess water of these reservoirs is released to the nearby lowlands to maintain the reservoirs' adequate water volume and pressure. Therefore, the water extents monitored using the remotely sensed data could show a minimal change with respect to time.

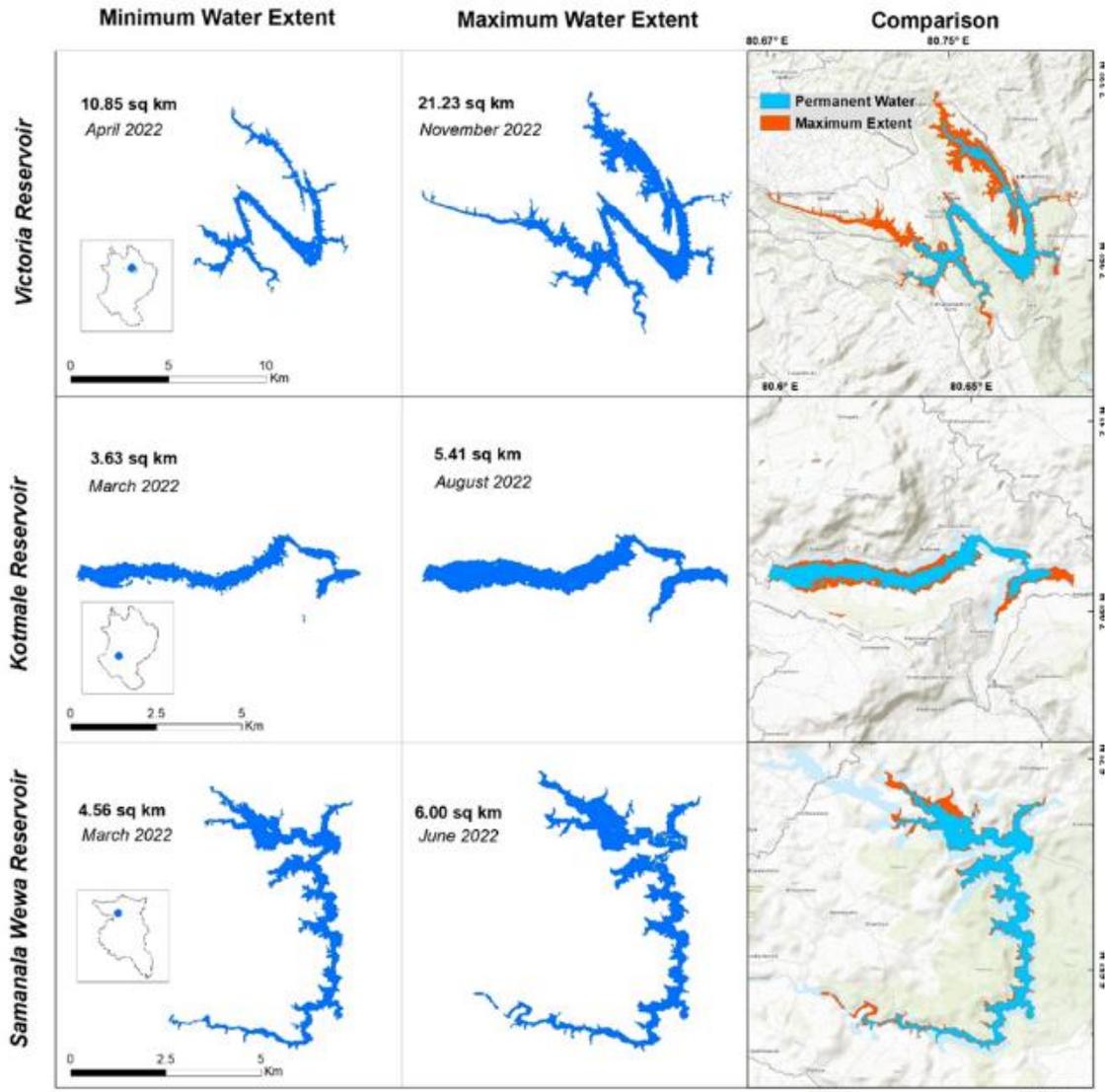

**Fig 8**. Maximum and Minimum Reservoir Extent Maps for Sri Lanka

The dynamic monitoring of reservoir water extent can support the researchers with the decision-making regarding protecting reservoir boundaries without harming the associative ecosystems and protecting the water quality, conserving the biodiversity, regulating runoff etc.[17]. The sustainability goals can be achieved with effective and robust water management policies, which could be developed after periodical monitoring of existing hydroelectric reservoirs.

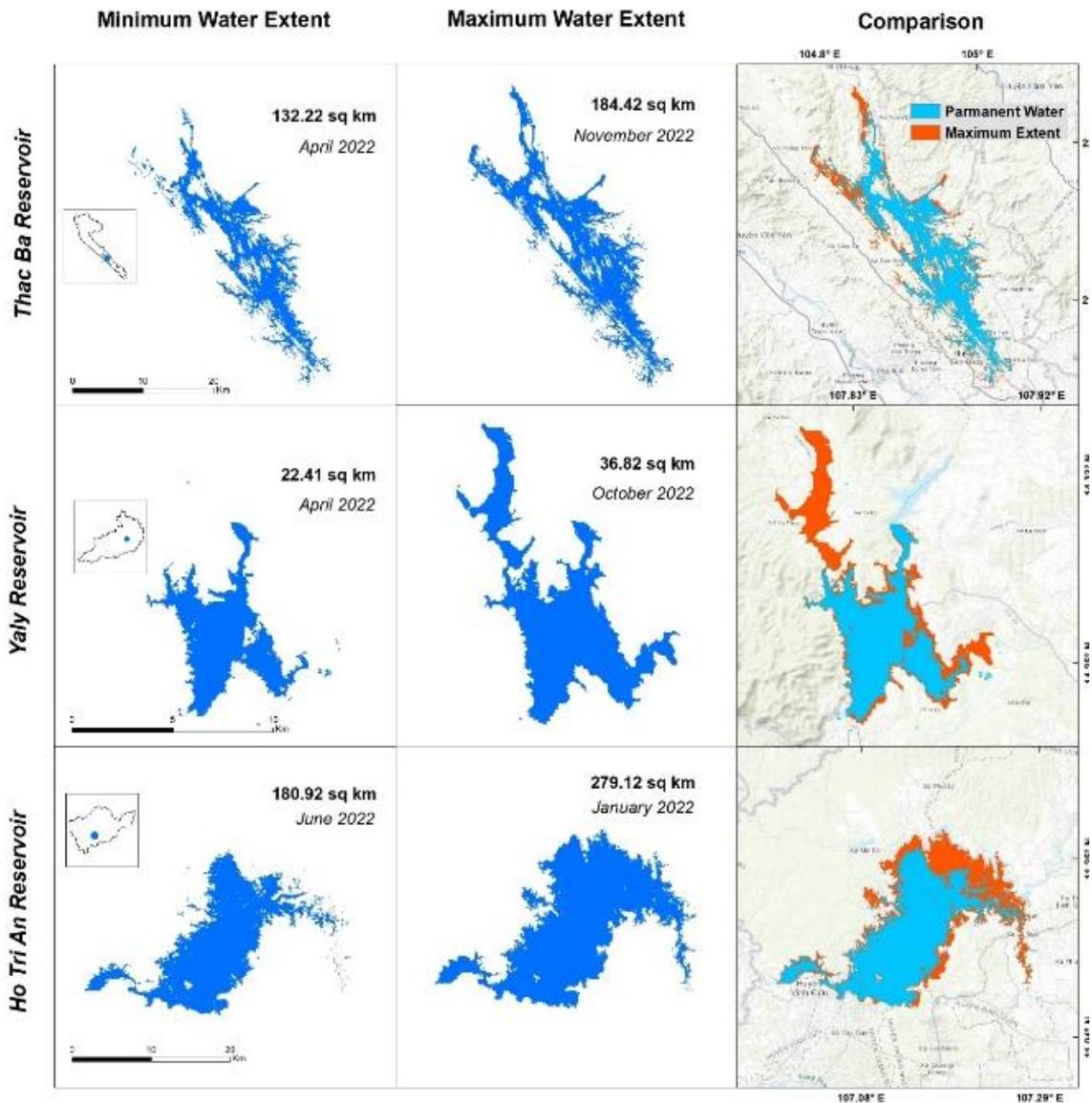

**Fig 9**. Maximum and Minimum Reservoir Extent Maps of Vietnam

## 5. CONCLUSION

The study focused on analyzing the rainfall patterns and reservoir water extents using cloud platforms for the two tropical monsoon countries, Vietnam and Sri Lanka. The study utilized remote sensing data, including Sentinel 1 SAR imagery, to assess the monthly water extent of selected hydroelectric reservoirs from 2017 to 2022. The results indicated that both Vietnam and Sri Lanka experience high annual average rainfall, with variations depending on the monsoon seasons. The analysis of rainfall patterns revealed higher precipitation during the monsoon period compared to the dry season. The box and whisker plots demonstrated the relationship between rainfall and reservoir water extents, showcasing the rise in water levels during and after the monsoon season.

The study provided valuable insights into the dynamics of reservoir water extents and their correlation with rainfall trends. The change of reservoir water extent follows the monsoon precipitation pattern.

Understanding these relationships is crucial for effective water resource management, particularly in the context of hydropower generation, flood management, and irrigation in both countries.

ACKNOWLEDGEMENT

This publication has been prepared as an output of the CGIAR Research Initiative on Low Emission Food Systems (Mitigate+). We would like to thank all funders who supported this research through their contributions to the CGIAR Trust Fund: https://www.cgiar.org/funders/

REFERENCES

[1] L. Pulvirenti, G. Squicciarino, E. Fiori, L. Ferraris, and S. Puca, "A Tool for Pre-Operational Daily Mapping of Floods and Permanent Water Using Sentinel-1 Data,"Remote Sens (Basel), vol. 13, no. 7, p. 1342, Apr. 2021, doi: 10.3390/rs13071342.

[2] A. B. Pal, D. Khare, P. K. Mishra, and L. Singh, "trend analysis of rainfall, temperature and runoff data: a case study of rangoon watershed in nepal,"International Journal of Students' Research in Technology & Management, vol. 5, no. 3, pp. 21–38, Nov. 2017, doi: 10.18510/ijsrtm.2017.535.

[3] H. Cong Vu, H. Le, and T. N. To, "Improvement of Multi-Purpose Reservoirs Operation: A Case Study in TraKhuc River Basin, Vietnam," 2021.

[4] Open Development Vietnam, "Water resources," Jan. 09, 2018. https://vietnam.opendevelopmentmekong.net/topics/water/ (accessed Jun. 24, 2023).

[5] Ministry of Power and Energy and Ceylon Electricity Board, "Feasibility Study for Expansion of Victoria Hydropower Station in Sri Lanka Final Report (Summary)," 2009.

[6] Ceylon Electricity Board, "Victoria Power Plant | Mahaweli Complex," 2023. https://www.mahawelicomplex.lk/victoria-power-plant/ (accessed Jun. 24, 2023).

[7] Ceylon Electricity Board, "Kotmale Power Plant | Mahaweli Complex," 2023. https://www.mahawelicomplex.lk/kotmale-power-plant/ (accessed Jun. 24, 2023).

[8] Ceylon Electricity Board, "CEB-GENERATION-NETWORK-2023".

[9] C. Kuenzer, I. Klein, T. Ullmann, E. Georgiou, R. Baumhauer, and S. Dech, "Remote Sensing of River Delta Inundation: Exploiting the Potential of Coarse Spatial Resolution, Temporally-Dense MODIS Time Series,"Remote Sens (Basel), vol. 7, no. 7, pp. 8516–8542, Jul. 2015, doi: 10.3390/rs70708516.

[10] R. B. L. Cavalcante, D. B. da S. Ferreira, P. R. M. Pontes, R. G. Tedeschi, C. P. W. da Costa, and E. B. de Souza, "Evaluation of extreme rainfall indices from CHIRPS precipitation estimates over the Brazilian Amazonia,"Atmos Res, vol. 238, p. 104879, Jul. 2020, doi: 10.1016/j.atmosres.2020.104879.

[11] Q. Zhao, L. Yu, X. Li, D. Peng, Y. Zhang, and P. Gong, "Progress and Trends in the Application of Google Earth and Google Earth Engine,"Remote Sens (Basel), vol. 13, no. 18, p. 3778, Sep. 2021, doi: 10.3390/rs13183778.

[12] Vietnam Government Portal, "overview on vietnam geography." https://vietnam.gov.vn/geography-68963 (accessed Jun. 24, 2023).

[13] International Hydropower Association, "Country Profile - Sri Lanka," 2022. https://www.hydropower.org/country-profiles/sri-lanka (accessed Jun. 24, 2023).

[14] K. H. Tran, M. Menenti, and L. Jia, "Surface Water Mapping and Flood Monitoring in the Mekong Delta Using Sentinel-1 SAR Time Series and Otsu Threshold,"Remote Sens (Basel), vol. 14, no. 22, p. 5721, Nov. 2022, doi: 10.3390/rs14225721.

[15] M. Giuliani, J. R. Lamontagne, P. M. Reed, and A. Castelletti, "A State-of-the-Art Review of Optimal Reservoir Control for Managing Conflicting Demands in a Changing World,"Water Resour Res, vol. 57, no. 12, Dec. 2021, doi: 10.1029/2021WR029927.

[16] S. Ghosh, and J. Mukherjee. "Earth observation data to strengthen flood resilience: a recent experience from the Irrawaddy River," Nat Haz, vol. 115, no. 3, pp 2749-2754, Feb. 2023, doi: 10.1007/s11069-022-05644-w.

[17] Q. Tang, B. Fu, A. L. Collins, A. Wen, X. He, and Y. Bao, "Developing a sustainable strategy to conserve reservoir marginal landscapes,"Natl Sci Rev, vol. 5, no. 1, pp. 10–14, Jan. 2018, doi: 10.1093/nsr/nwx102.